# Multilevel Text Normalization with Sequence-to-Sequence Networks and Multisource Learning

**Tatyana Ruzsics**  **Tanja Samardžić**
Text Group, URPP Language and Space,
University of Zurich, Switzerland
tatiana.ruzsics, tanja.samardzic@uzh.ch

## Abstract

We define multilevel text normalization as sequence-to-sequence processing that transforms naturally noisy text into a sequence of normalized units of meaning (morphemes) in three steps: 1) writing normalization, 2) lemmatization, 3) canonical segmentation. These steps are traditionally considered separate NLP tasks, with diverse solutions, evaluation schemes and data sources. We exploit the fact that all these tasks involve sub-word sequence-to-sequence transformation to propose a systematic solution for all of them using neural encoder-decoder technology. The specific challenge that we tackle in this paper is integrating the traditional know-how on separate tasks into the neural sequence-to-sequence framework to improve the state of the art. We address this challenge by enriching the general framework with mechanisms that allow processing the information on multiple levels of text organization (characters, morphemes, words, sentences) in combination with structural information (multilevel language model, part-of-speech) and heterogeneous sources (text, dictionaries). We show that our solution consistently improves on the current methods in all three steps. In addition, we analyze the performance of our system to show the specific contribution of the integrating components to the overall improvement.

## 1 Introduction

Various kinds of texts are increasingly seen as a vast source of knowledge that can be structured and extracted using natural language processing (NLP). An important problem that needs to be solved in initial steps of the NLP pipeline is the fact that texts are noisy: the same units of meaning (words or morphemes) can appear in various, often unpredictable, shapes. Consider, for instance, the English word *general* in Figure 1. It can appear as an individual word, or as a part of other words (*generalize*, *generalization*, *generally*). These words in turn can appear in different forms, such as tenses of the verb *generalize* (*generalized*, *generalizes*, *generalizing*). Finally, each of these tenses can be further varied due to different (or lacking) writing standards, typos, abbreviations etc. (e.g. *genrealizes*, *generalises*, *g-izes*). It is crucial for automatic text processing to recognize all these variants as instances of the same word (or morpheme).

Traditionally, each of these levels of variation constituted a separate NLP task. The goal of *writing normalization* has been to identify the same word types in different writings. The task of *lemmatization* has been to deal with morphological inflection, and *canonical segmentation* has been concerned with identifying the same morphemes in different words. The first three columns in Figure 1 illustrate this division. These tasks have been performed using different strategies, algorithms, data sets and evaluation schemes. Recently, it has been recognized that all these tasks involve sequence-to-sequence processing, which can be solved with the neural sequence-to-sequence (seq2seq) technology (Cho et al., 2014; Sutskever et al., 2014) leading to improvements over traditional methods. These improvements, however, remain bound to the separate tasks, not fully exploiting the available data and techniques.

In this paper, we approach all the three tasks with a single, configurable seq2seq framework exploiting a wide range of the available data. In addition to the usual seq2seq training sets, which consist of pairs *(variant, canonical form)*, we make use of pretrained language models at different levels (character, morpheme, word), different kinds of context information including part-of-speech (POS) tagging, and external dictionaries. We show how the complementary information in these different sources allows consistent further improve-ments over the most up-to-date seq2seq solutions. To better understand the contribution of our specific solutions, we analyze the performance of our system breaking down the test cases into specific categories (e.g. ambiguous, out-of-vocabulary items).

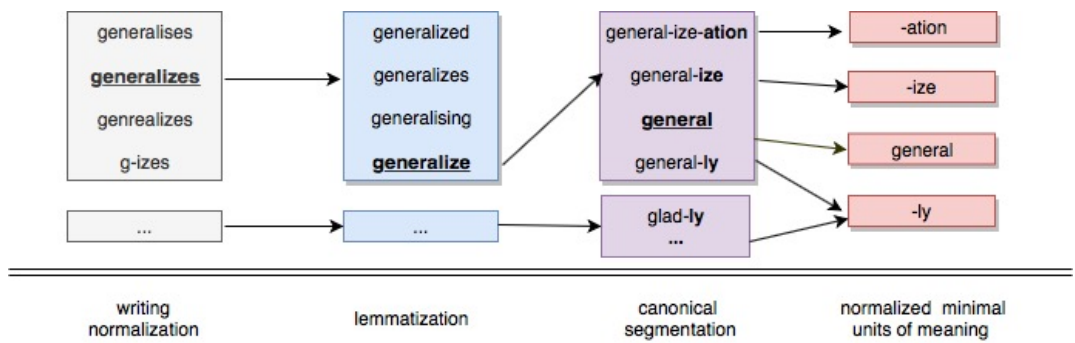

Figure 1: From noisy to normalized text: several forms are mapped to one (canonical) at each level. The process is repeated till there is no variation.

## 2 Methods

Our approach relies on the fact that all three kinds of reducing variants to a canonical form illustrated in Figure 1 can be commonly formalized as a transformation of a source sequence of characters into a target sequence of characters and solved with character-level seq2seq methods. More precisely, we build our system on top of a neural machine translation (NMT) system with attention (Bahdanau et al., 2014) outlined in Section 2.1. In plain NMT, only the sequence that is transformed (a word in our case) is considered. Our modifications of the plain NMT framework allow incorporating additional sequential context signal that can be found on multiple structural levels of both the source and the target sequence. We hence distinguish between *sequential source context* and *sequential target context*, which we exploit in different ways for different purposes.

The idea behind the modifications introduced to include additional **source** context is to deal with word-level ambiguity, traditionally addressed only in the task of lemmatization, where POS tags are commonly given as input together with source-side word. For example, POS tags can help to distinguish between normalizing a Russian source word *всего* into words *всё* 'all' (pronoun), *весь* 'whole' (determiner) and *всего* 'just' (adverb). However, POS tags do not always provide enough information to resolve the ambiguity. To illustrate, Russian input adjective *большую* can be normalized as *большой* 'big' or *больший* 'the biggest'.

One way to summarize the sequential source context is to use a (manually annotated) POS tag of the input word. Another possibility is to include the surrounding words as features, which is currently done with a hierarchical bi-LSTM and proved to work well in lemmatization (Chakrabarty et al., 2017; Kondratyuk et al., 2018) and tagging tasks (Ling et al., 2015; Plank et al., 2016; Yasunaga et al., 2018a). In the latter approach, a lower-level bi-LSTM encodes the surrounding words on a character level and a higher-level LSTM summarizes the entire context by reading the resulting sequences of the character embeddings to the left and to the right of the input word. In this way, both character-level and word-level signals from the source sequential context are included.

We make use of both possibilities. We first consider two separate paths, NMT+POS and NMT+context, where the two kinds of the context information are given separately as input to the NMT decoder. Next, we combine the two types of the source context in a flat manner (NMT+context+Gold POS) and in a hierarchical fashion (NMT+context+Predicted POS). The latter model first uses context to predict a POS tag, then uses the context together with the predicted POS tag for the task prediction. The details of the architectures integrating the source context are described in Section 2.2.

Our proposed source-context models are expected to reveal 1) to which extent automatic encoding of the context with hierarchical bi-LSTM can replace manual POS-tags 2) complementarity of the two ways to encode the context in targeting the cases where POS-tag is not enough to resolve ambiguity.

The sequential **target** context is part of plain NMT, but, again, only within the sequence that is transformed and on the level at which the transformation is performed (character in our case). However, there is additional target-side information that can help in our task. In the task of canonical segmentation, for example, a morpheme-level language model can help normalize English *incessant* into a sequence of canonical segments *in cease ant*. In the task of writing normalization, one source-side word can often be normalized with a sequence of multiple words, such as for example, a source Swiss dialect word *kömmer* 'we can' normalized as a sequence of Standard German words *können wir*.

In order to include additional target-side sequential context we follow the line of the work, where pretrained target-side language model (LM) is integrated into a seq2seq framework for speech recognition (Bahdanau et al., 2016; Chorowski and Jaitly, 2016; Kannan et al., 2018) and machine translation (Gülçehre et al., 2015; Gulcehre et al., 2017). More specifically, we follow the multilevel LM integration approach of Ruzsics and Samardžić (2017) to fuse, in a log-linear fashion, the target-side *higher level language model* (HLLM) with the plain NMT scores during the inference stage. The scores are combined only at the higher-level units boundaries, using a two-level beam decoding approach described in more detail in Section 2.3. We expect that HLLM over the target data specifically developed for the task (manual normalizations in training pairs) will correct character-level NMT for the segments that can be found in the training set. In addition, we expect that the performance can be further improved by consulting non-adapted, noisy target source data. In this sense, one can employ unsegmented dictionaries or unaligned raw text to deal with the items that do not appear in the training set. We therefore work with such target-side sources too.

### 2.1 NMT base

First, we describe the basic configuration of the NMT system, an encoder-decoder model with soft attention (Bahdanau et al., 2014; Luong et al., 2015), that we use for all our neural experiments. In order to formalize our task, we define two vocabulary sets, $\Sigma$ consisting of the character symbols that form the source sequences and $\Sigma_n$ of the character symbols that form the target sequences. Then, our task is to learn a mapping from an original character sequence $x \in \Sigma^*$ to its normalized form $y \in \Sigma_n^*$.

The model transforms the input sequence $x = (x_1, \ldots, x_{n_x})$ into a sequence of hidden states $\mathbf{h}^x = (\mathbf{h}_1^x, \ldots, \mathbf{h}_{n_x}^x)$ using a bidirectional LSTM encoder:

$$\mathbf{h}^x = [\overrightarrow{f}(\mathbf{x}_{1:n_x}); \overleftarrow{f}(\mathbf{x}_{n_x:1})] \quad (1)$$

where (forward) character representation of $x_i$ is

$$\overrightarrow{f}(\mathbf{x}_{1:n_x})_i = f(\mathbf{x}_i, \overrightarrow{f}(\mathbf{x}_{1:n_x})_{i-1}) \quad (2)$$

and f stands for an LSTM cell (Hochreiter and Schmidhuber, 1997). The decoder LSTM transforms the internal fixed-length input representation $\mathbf{h}_x$ into a variable length output sequence $y = (y_1, \ldots, y_{n_y})$. At each prediction step $t$, the decoder reads the previous output $\mathbf{y}_{t-1}$ and outputs a hidden state representation $\mathbf{s}_t$:

$$\mathbf{s}_t = f(\mathbf{s}_{t-1}, \mathbf{y}_{t-1}), \quad t = 1, \ldots, n_y \quad (3)$$

The conditional probability over output characters is modeled at each prediction step $t$ as a function of the current decoder hidden state $\mathbf{s}_t$ and the current context vector $\mathbf{c}_t$:

$$p(y_t|y_1, \ldots, y_{t-1}, x) = g(\mathbf{s}_t, \mathbf{c}_t) \quad (4)$$

where $g$ is a concatenation layer followed by a softmax layer (Luong et al., 2015). The context vector $\mathbf{c}_t$ is computed at each step from the encoded input as a weighted sum of the hidden states using attention model described in (Luong et al., 2015).

The training objective is to maximize the conditional log-likelihood of the training corpus:

$$L = \frac{1}{N} \sum_{(x,y)} \sum_{t=1}^{n_y} \log p(y_t|y_1, \ldots, y_{t-1}, x) \quad (5)$$

where $N$ is the number of training pairs $(x, y)$.

### 2.2 Integrating source context

We consider several architectures for incorporating source contextual information into the plain NMT system. In particular, each of these methods incorporates contextual features associated with the whole input word $x$ into each character-level prediction step 4 of the NMT model.

**NMT+context** In our first system, we assume that the input to the system is a pair of a source word $x$ together with its context $s = (s_1, \ldots, s_{i-1}, x, s_{i+1}, \ldots s_{n_s})$, where $x = s_i$. For example, the context can be represented by a sentence in text or an utterance in speech.

We encode the context $s$ for the word $x = s_i$ using a hierarchical bi-LSTM architecture. First, a lower-level bi-LSTM encodes separately all the words in the sentence $s$ on the character-level into a sequence of lower hidden states and each word is then represented by its character embedding consisting of the last forward and last backward lower-level LSTM states: $\mathbf{e}^s = \mathbf{e}_1^s, \ldots, \mathbf{e}_{n_s}^s$:

$$\mathbf{e}_j^s = [\overrightarrow{f}(\mathbf{c}_{1:n_{s_j}})_{n_{s_j}}; \overleftarrow{f}(\mathbf{c}_{n_{s_j}:1})_{n_{s_j}}] \quad (6)$$

where $s_j = c_1, \ldots, c_{n_{s_j}}, c_k \in \Sigma$ is a character sequence of the context word $s_j$.

A higher-level bi-LSTM encodes a sequence of the character representations of the context words $\mathbf{e}^s$ into a sequence of higher hidden states $\mathbf{H}^s$:

$$\mathbf{H}_j^s = [\overrightarrow{f}(\mathbf{e}_{1:n_s}^s)_j; \overleftarrow{f}(\mathbf{e}_{n_s:1}^s)_j] \quad (7)$$

We adapt the plain NMT system to feed the context representation $H^x = H_i^s$ of the input word $x$, together with the current decoder hidden state $\mathbf{s}_t$ and the current context vector $\mathbf{c}_t$, in order to predict the next output character as follows:

$$p(y_t|y_1, \ldots, y_{t-1}, x) = g(\mathbf{s}_t, \mathbf{c}_t, \mathbf{H}^x) \quad (8)$$

**NMT+Gold POS** In the second system, we assume that the contextual input is represented only by a POS-tag of the input word. In addition to the two vocabularies that contain the source $\Sigma$ and target $\Sigma_n$ characters, we consider a vocabulary of the possible POS tags $\Sigma_f$. Our task in this setting is to learn a mapping from an input pair $(x, f_x = f_1 + \ldots + f_k)$ of a source character sequence $x \in \Sigma^*$ and its POS feature $f_x$ (possibly consisting of one or more tags $f_i \in \Sigma_f$) to its target normalized form $y \in \Sigma_n^*$. We embed the POS tags $f_i \in \Sigma_f$ into their vector representations $\mathbf{f}_i$, which are learned by the system. In cases where the feature input is a composition, i.e. it consists of several POS tags $f_1 + \ldots + f_k$[1], we use an average of the corresponding vector embeddings $(\mathbf{f}_1 + \ldots + \mathbf{f}_k)/n$ as representation. We modify the prediction stage of the plain NMT system by feeding the POS feature at each prediction step:

$$p(y_t|y_1, \ldots, y_{t-1}, x) = g(\mathbf{s}_t, \mathbf{c}_t, \mathbf{f}_x) \quad (9)$$

**NMT+context+Gold_POS** The next system combines the context and POS-tag input together to explore their complementary relation. This system learns a mapping from an input pair $(x, f_x, s)$ of a source word $x \in \Sigma^*$, its POS feature $f_x \in \Sigma_f$ and context $s = (s_1, \ldots, s_{i-1}, x, s_{i+1}, \ldots s_{n_s})$ to its target normalized form $y \in \Sigma_n^*$. We use the methods from the previous systems to encode the input and modify the prediction stage as follows:

$$p(y_t|y_1, \ldots, y_{t-1}, x) = g(\mathbf{s}_t, \mathbf{c}_t, \mathbf{f}_x, \mathbf{H}^x) \quad (10)$$

**NMT+context+Predicted_POS** Finally, we propose a system which targets a more realistic setting where POS-features are not given at the test time. This system is trained to predict both the tag of the input and its normalized form. It thus learns a mapping from $(x, s)$ of a source word $x \in \Sigma^*$ and its context $s$ to the pair $(y, f_x)$ of the normalized form $y \in \Sigma_n^*$ and POS-tag $f_x$. The prediction step is sequential: the system first predicts the tag $f_x$ according to the distribution

$$p(f_x|x) = \mathsf{softmax}(W_f \mathbf{H}^x) \quad (11)$$

where $W_f$ is a learned parameter. In the next step, the system uses the predicted tag in predicting the normalized form as in (10). At the train time, we use the gold tags for the prediction of the normalized form. The parameters of the model are trained by maximizing a combined objective:

$$L = \frac{1}{N} \sum_{(x, f_x, s, y)} (\alpha \log p(f_x|x, s) + \sum_{t=1}^{n_y} \log p(y_t|y_1, \ldots, y_{t-1}, x, f_x, s)) \quad (12)$$

### 2.3 Integrating target context (NMT+HLLM)

Before the integration, we assume that an NMT model and a HLLM are trained separately. NMT is trained on character sequences while HLLM is trained over the higher level segments of the target data, i.e. morphemes or words. We combine the two models at the inference stage with a beam search that runs on two levels of granularity. First, it produces the output sequence hypotheses at the character level using NMT scores until the first integration point, where the beam-size number of best hypotheses end with a boundary symbol. The boundary symbol marks the end of a segment. At this step we re-score the normalization hypotheses with a weighted sum of the NMT score and the HLLM score. After that the process is continued till the next synchronization point until all the hypotheses in the beam are closed with an end-of-word symbol. In this way, the HLLM score helps to evaluate how probable the last generated segment is based on the predicted segment history, that is the sequence of segments generated at the previous integration time steps.

---
[1] Such cases of multiple POS tags are common in the task of writing normalization where one source word is aligned to multiple target words.

## 3 Related Work

Writing normalization problem has been initially addressed in historical text normalization with automatic induction of rules (Reffle, 2011; Bollmann et al., 2012) or similarity-based form matching inspired by spellchecking (Baron and Rayson, 2008; Pettersson et al., 2013). A major breakthrough in performing the task was achieved when it was approached as a case of character sequence transformation and tackled with character-level statistical machine translation (SMT) in computer-mediated communication (Clercq et al., 2013; Ljubesic et al., 2014), historical texts (Pettersson et al., 2014) and dialects (Samardžić et al., 2015; Scherrer and Ljubešić, 2016). With the introduction of neural seq2seq methods, there appear numerous efforts to apply this paradigm to the normalization of historical texts (Bollmann and Søgaard, 2016; Bollmann et al., 2017; Korchagina, 2017; Tang et al., 2018; Robertson and Goldwater, 2018) and dialect data (Honnet et al., 2018; Lusetti et al., 2018). While additional data on the target side can be easily integrated into the SMT framework (Scherrer and Ljubešić, 2016; Honnet et al., 2018), such augmentation of the neural systems has been less explored: Bollmann et al. (2017) apply lexicon-checking during decoding while Lusetti et al. (2018) work with two-level decoding with HLLM trained on an additional target-side corpus. We follow the latter approach but apply it to more noisy target data, not specifically developed for to the task. Also, the employment of source context was only studied in the SMT approach of Scherrer and Ljubešić (2016) while we are not aware of such modifications for NMT systems in the task of writing normalization.

For lemmatization systems, we observe a similar shift in the paradigm from rule/heuristics based approaches (Koskenniemi, 1984; Plisson et al., 2004) to the supervised learning methods which treat lemmatization either as edit scripts or edit trees classification problem (Chrupala et al., 2008; Müller et al., 2015; Chakrabarty et al., 2017), prefix and suffix transformation problem (Jursic et al., 2010; Gesmundo and Samardzic, 2012) or character-level string transduction process (Dreyer et al., 2008; Nicolai and Kondrak, 2016; Eger, 2015). The last line of research was recently extended with the neural seq2seq systems (Bergmanis and Goldwater, 2018; Kondratyuk et al., 2018). Concerning the incorporation of context into NMT systems, it was shown that inclusion of raw character-level source context into NMT systems (Bergmanis and Goldwater, 2018) can help in languages with higher ambiguity, with the results comparable to previous context-sensitive lemmatizers which use context in the form of POS tags (Chrupala et al., 2008), richer morphological tags (Müller et al., 2015) or encoded with hierarchical LSTMs (Chakrabarty et al., 2017). On the other hand, Kondratyuk et al. (2018) showed that a context-aware NMT system augmented with an auxiliary task of predicting morphological tags achieves state-of-the-art performance in lemmatization of morphologically-rich Czech, German, and Arabic. However, it has not been studied to which extent two types of the source context signal, a raw sequence of characters/words and POS/morphological tags, can be mutually substituted and to which extent they are complementary. We focus on this analysis by testing a single neural systems which can include these signals separately or fuse them together. We analyze our performance across 20 languages and identify the complementary aspects of the two types of context encoding.

Regarding the task of canonical segmentation, i.e. segmenting words into canonical segments with recognition of orthographic changes that take place during word formation, we underline its difference from the task of surface segmentation (segmenting words into substrings) which has its own long tradition of methods. Initially, canonical segmentation was approached with unsupervised methods (Dasgupta and Ng, 2006; Naradowsky and Goldwater, 2009), whereas later, the task was casted as a sequence transduction problem and tackled with conditional random fields

(Cotterell et al., 2016; Cotterell and Schütze, 2018) and neural seq2seq methods (Kann et al., 2016; Ruzsics and Samardžić, 2017). All supervised systems make use of additional heterogeneous target data except the work of Ruzsics and Samardžić (2017) which exploits the signal from higher level units in the target side of the parallel train data. We extend their idea in our target-context approach by incorporating higher level units signal from additional lexical resources into the NMT method and show that such approach leads to the improvement over the previous state-of-the-art canonical segmenter of Cotterell and Schütze (2018).

## 4 Data and Experimental setup

We evaluate our approach by comparing the performance of our system to previous solutions for all the three tasks.

**Writing Normalization** For the text normalization experiments we use the ArchiMob corpus (Samardzic et al., 2016) which represents German linguistic varieties spoken within the territory of Switzerland. In order to reduce intra-speaker and regional variation in transcriptions, each original word form is manually annotated with a normalized form in a subset of 8 recordings which we use in our experiments. Out of 8 documents, 6 are manually annotated with POS tags while the remaining 2 are tagged with a CRF tagger. The utterances in the corpus are split into syntactically and prosodically motivated segments of 4-8 seconds. We use these segments to extract context information in our source-context experiments. The full dataset (8 documents) is split into train (12,087 segments with 94,122 words), development (1,459 segments with 12,197 words) and test sets (1,055 segments with 8,124 words). For the target-context experiment we use the Standard German OpenSubtitles2016 corpus (Lison et al., 2018), 185M tokens in size.

We apply our source-context methods and repeat these experiment in a combination with the target-context method (NMT+HLLM) where we train a word-level LM on the Open Subtitles corpus. We compare our results directly with the current state-of-the-art model of Scherrer and Ljubešić (2016).[2]

[2] The results for this model were published for a smaller manually annotated portion of Archimob corpus available at that time. We rerun their model on the extended dataset.

**Lemmatization** We use the same data as in the lemmatization experiments of Bergmanis and Goldwater (2018), i.e. standard splits of the Universal Dependency Treebank (UDT) v2.0[3] (Nivre et al., 2017) datasets for 20 languages. The data varies in size among the languages, for example, train data ranges from 16K tokens for Hungarian to 260K tokens for Hindi. In the source-context experiments we extract context within sentence boundaries and use universal POS tags from the corpus. We run experiments with our source-context methods and show the results of the Lematus system from (Bergmanis and Goldwater, 2018) for comparison.

**Canonical Segmentation** We use English and German datasplits for canonical segmentation described in Cotterell and Schütze (2018)[4]. Both sets consist of 10,000 word types split randomly 10 times into 8,000 train, 1,000 development and 1,000 test pairs. We report the numbers on 5 splits and 10 splits to compare to the results of the existing models. As an additional target data for our target-context experiments we use ASPELL dictionaries[5], which amount to 120K word types for English and 365K word types for German.

We apply the target-context method (NMT+HLLM) where we train a target-side LM over morpheme sequences in the original train target data augmented with additional target data from ASPELL. To analyze how our target-context method performs on rare word types, we also add the results for plain NMT model and NMT+HLLM model where HLLM is trained only over morpheme sequences in the original train target data. For direct comparison, we show the results of the neural reranker model of Kann et al. (2016), joint transduction and segmentation model of Cotterell et al. (2016). We also show the results of the joint transduction and segmentation model with word embeddings (trained on Wikipedia data) of Cotterell and Schütze (2018) which is not directly comparable to previous methods due to use of different additional data.

**Hyperparameters** We use the same hyperparameters across the tasks unless mentioned otherwise. The character embeddings are shared be-tween input (source) and output (target) vocabulary and are set to 100. The size of POS embeddings is 50. All LSTM networks have 200 hidden units. We apply an ensemble of 5 NMT models, where each model is trained using SGD optimization. The models are trained for a maximum of 30 epochs (except writing normalization task where we apply 40 epochs), possibly stopping earlier if the performance measured on the development set is not improving after 10 epochs. The training examples (sentences in case of source-context models) are shuffled before each epoch. The contribution $\alpha$ of POS-tagging loss in the training objective (12) is set to 0.2. We use beam decoding with size 3. In experiments with target-context, we train 3-gram statistical HLLM and tune the weights of NMT and HLLM used at inference with MERT optimization by maximizing accuracy score on development data.

[3] http://hdl.handle.net/11234/1-1983
[4] http://ryancotterell.github.io/canonical-segmentation/
[5] http://cistern.cis.lmu.de/chipmunk/supplement.tar.gz

**POS Baseline** In the settings with source context input, we apply a simple baseline which addresses ambiguous words and quantifies the difficulty of their normalization in the presence of POS tags input. To this end, we consider three classes of input words in the test set: *New*, *Ambiguous* and *Unique*. The *New* category includes the words that have not been observed in the training set which the baseline simply copies. The *Unique* words are associated with exactly one target-side form (normalization) in the train set, which is selected by the baseline at test time. The last category, *Ambiguous*, consists of input words which are associated with more than one target-side candidate in the train set. We consider two subclasses of the *Ambiguous* words. The first subclass, *POS-unambiguous*, consists of words for which each POS tag that appears together with this word in the train set can be associated with a unique target-side form, i.e. there is a unique pair (target word, POS tag). This form is then selected by the baseline at the test time. The other subclass, *POS-ambiguous*, consists of the words whose at least one normalization has more than one POS tag in the train set. In such cases, if the tag of the test input word has not been observed with this word at the train time, the baseline selects the most frequent target of this word in the train set. Otherwise, it selects the most frequent prediction corresponding to the test input pair (word, POS tag) or a random form out of all its possible normalizations in case of a tie.

## 5 Results and Discussion

**Writing Normalization** The results of the experiments on writing normalization task are shown in Table 1. We observe that source-context models improve the results of the previous state-of-the-art model from (Scherrer and Ljubešić, 2016). Adding target-context component pushes the results further resulting in 92.43% overall accuracy for the best performing NMT+context+Gold POS model.

Comparing the results with and without additional target-context component (upper and lower half of the Table 1) we note that all the models benefit from the additional target data in the category of *New* words resulting in 3-4% improvement within this subcategory across the models. Examining the results for *Ambiguous* words in the lower half of the table, we note that in comparison to the plain NMT system, adding raw context signal (NMT+context) targets *POS-amb.* category while POS tags (NMT+Gold POS) help with the *POS-unamb.* words. Overall, there is a stronger preference for the former model due to the dominance of *POS-amb.* subcategory as well as a high weight of *Ambiguous* words in the data. Our baseline explains well the difficulty of the task and the behaviour of the NMT+Gold POS model: both reach an accuracy of 88% on the *POS-unamb.* subcategory. NMT+context improves further the baseline accuracy on the *POS-amb.* category by around 6%. Combining the two source context signals in a flat architecture (NMT+context+Gold POS) brings further small improvements in each category compared to the individual models. It is closely followed by the hierarchical architecture (NMT+context+Predicted POS) which interestingly works better on *POS-amb.*. We hypothesize that this is due to the presence of non-gold tags in the subset of the data. While predicting POS tags within the hierarchical model helps on the *POS-unamb.* category (+1.5% points) compared to the pure context model (NMT+context), it is still lower by 5% points in overall accuracy in comparison to the use of gold POS tags in NMT+Gold POS model and the baseline.

**Lemmatization** The results of the lemmatization experiments averaged over 20 languages are presented in Table 2. Among our source-context models, the best overall result of 96.68% is achieved by NMT+context+Gold POS model. As in the normalization results, we note that the two types of the source context signals work complementary to each other: NMT+context model improves performance of the POS-baseline model by around 3% points on the *POS-amb.* category while inclusion of POS signal (NMT+Gold POS) results in the comparable to the baseline accuracy on the *POS-unamb.* category. Also similarly to the normalization task, though inclusion of POS prediction step into the hierarchical model helps on the *POS-unamb.* category (95.29% versus 94.23% for NMT+context), it is inferior in comparison to the use of gold POS tags on this category resulting in 99.29% accuracy for NMT+context+Gold POS accuracy and 99.35% for NMT+Gold POS models. We conclude that the use of POS tags signal is helpful when the POS prediction step captures the tags reliably.

| | No of, % | POS-baseline | NMT | NMT + context | NMT + context + Gold POS | NMT + context + Predicted POS | NMT + Gold POS | SMT + context* |
|---|---|---|---|---|---|---|---|---|
| Archimob: | | | | | | | | |
| Total | | 83.43 | 88.39 | 91.85 | 91.99 | **92.02** | 90.10 | - |
| Unamb. | 42.93 | 98.71 | 98.68 | 98.74 | **98.80** | 98.74 | 98.60 | - |
| New | 10.60 | 8.13 | 60.74 | **62.14** | 61.67 | 61.32 | 61.67 | - |
| Amb. | 46.47 | 86.49 | 85.19 | 92.26 | 92.61 | **92.82** | 88.74 | - |
| Archimob + OpenSubt (+HLLM): | | | | | | | | |
| Total | | 83.43 | 88.69 | 92.18 | **92.43** | 92.31 | 90.46 | 89.73 |
| Unamb. | 42.93 | 98.71 | 98.68 | 98.74 | **98.82** | 98.71 | 98.65 | - |
| New | 10.60 | 8.13 | 63.30 | 65.39 | **66.09** | 64.46 | 64.92 | - |
| Amb. | 46.47 | 86.49 | 85.25 | 92.24 | 92.53 | **92.74** | 88.72 | - |
| POS-unamb. | 5.30 | 88.00 | 80.00 | 84.00 | **89.00** | 85.50 | 88.00 | - |
| POS-amb. | 94.70 | 86.41 | 85.54 | 92.70 | 92.73 | **93.15** | 88.76 | - |

Table 1: Performance on the task of Swiss German normalization by data categories (Word accuracy). The data categories are described in Section 4. *SMT+context is a model of Scherrer and Ljubešić (2016).

While there is a strong preference for the NMT+context+Gold POS model across all languages, it is closely followed by the simpler NMT+Gold POS model. This brings up the question, to which extent adding context to the latter model is advantageous for the task of lemmatization language-wise. By plotting the difference in accuracy of the two models across languages in Figure 2, we observe that the context preference is highly visible in two languages: Hindi and Urdu. This can be explained by the distribution of test cases over the analyzed subcategories. If we select the languages where the proportion of *Ambiguous* category prevails over the *New* category (Figure 3) and look at the composition of the *Ambiguous* category by plotting *POS-amb./POS-unamb.* ratio (Figure 4), we find Hindu and Urdu ranked on the top.

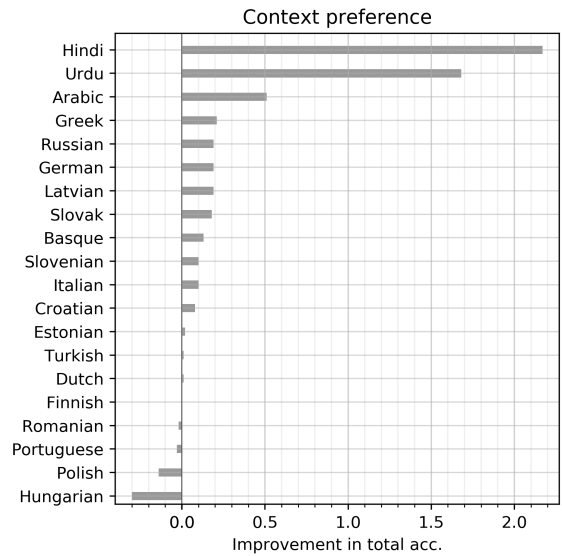

Figure 2: Context preference across languages in the lemmatization task quantified as difference in accuracy between NMT+context+Gold POS and NMT+Gold POS models.

Most of the other languages show a moderate preference for the context (less than 0.5% difference) which becomes much more remarkable if we zoom into the *POS-amb.* category and plot context preference (Figure 5), which is quantified as the difference in accuracy between NMT+context+Gold POS and NMT+Gold POS models on this subcategory. This gain shrinks on the total scale due to a low weight of this subcategory among the languages.

|  | POS-baseline | NMT + context | NMT + context + Gold POS | NMT + context + Predicted POS | NMT + Gold POS | Lematus* |
|---|---|---|---|---|---|---|
| *Total* | 85.57 | 95.77 | **97.18** | 95.90 | 96.96 | 94.9 |
| *Unamb.* | 99.25 | 99.27 | 99.57 | 99.15 | **99.57** | - |
| *New* | 40.41 | 85.89 | 90.18 | 86.38 | **90.39** | - |
| *Amb.* | 95.20 | 93.44 | **96.91** | 94.56 | 95.58 | - |
| *POS-unamb.* | 99.30 | 94.24 | 99.29 | 95.29 | **99.35** | - |
| *POS-amb.* | 84.95 | 88.37 | **90.27** | 89.55 | 85.73 | - |

Table 2: Performance on the task of lemmatization (Word accuracy averaged over 20 languages). Lematus* - context-sensitive model of Bergmanis and Goldwater (2018). The data categories are described in Section 4.

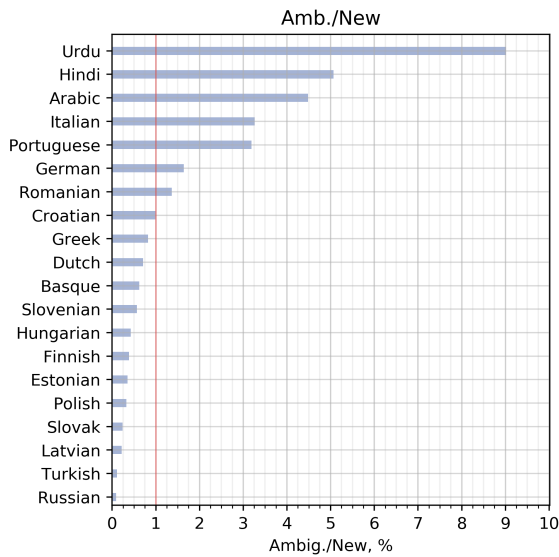

Figure 3: Subcategories structure in the lemmatization data: Ambiguous vs. New ratio

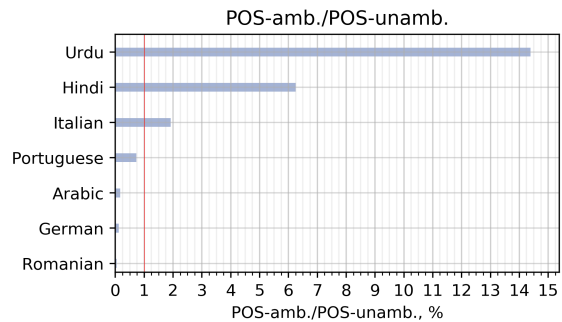

Figure 4: Subcategories structure in the lemmatization data: Ambiguous POS-amb. vs. POS-unamb. ratio

**Canonical Morphological Segmentation** The results of the segmentation experiments are presented in Table 3. We observe that the target context model NMT+HLLM with additional ASPELL target data reaches 83% accuracy for both English and German improving over the other models which rely on the additional data as well the state-of-the-art Joint+Vec model which uses even extra data from Wikipedia.

To analyze the performance of the models on the *New* category we consider two subcategories of unseen input words[6]: *New morphemes* are the words with segmentation consisting of new morphemes, and *New combinations* are the words with segmentation consisting of a new combination of seen morphemes. While NMT+HLLM model with HLLM trained only on the target side of the parallel train data shows improvement over the plain NMT system in the *New combinations* category, the use of additional target data helps with the performance in the *New morphemes* category mainly due to unseen roots that can be found in additional data. While the latter looses performance on the *New combinations* compared to the former, overall it achieves a better result for both languages which can be explained by a high weight of the *New morphemes* category in the data structure.

## 6 Conclusion and Future Work

We have proposed a novel processing framework for a systematic approach to upstream string transformation tasks using character-level NMT. Our system is adaptable allowing processing the information from different sources and on multiple structural levels (characters, morphemes, words, sentences) on the source and target side, depending on what training data is available and useful for what task. Our experiments on writing normalization, lemmatization and canonical segmentation and/or regularization with adversarial training (Yasunaga et al., 2018b) or joint training with word rarity (Plank et al., 2016)), could be advantageous. On the other hand, one could investigate to which extent the consideration of more language-specific tags and/or morphological features which are often used in lemmatizers (Toutanova and Cherry, 2009; Müller et al., 2015; Kondratyuk et al., 2018) can complement the raw source context signal. Another direction of the future work could explore the use of additional data in the form of word representations (word2vec (Mikolov et al., 2013) or recently introduced ELMo (Peters et al., 2018)) which have been shown to make the systems more robust to change of domain (Eger et al., 2016). They can be straightforwardly incorporated to our framework as an additional input into the hierarchical bi-LSTM for the context encoding.

---
[6] Seen input words constitute less than 1.5% of the test data for English and less than 0.6% - for German, on average across splits, therefore we do not present the numbers on this category.

|              | No of, %    | NMT        | NMT+HLLM   | +ASPELL |         |            | +WIKI |
|              |             |            |            | RR*     | Joint*  | NMT+HLLM   | Joint+Vec* |
|---|---|---|---|---|---|---|---|
| English-5    |             | 80. (1.)   | 80. (1.)   | 81. (1.) |        | **82.** (1.) |            |
| English-10:  |             | 79. (1.42) | 80. (1.15) |         | 77. (1.3) | **83.** (1.32) | 82. (2.)   |
| *New morph.* | 69.3 (1.2)  | 80. (1.)   | 79. (1.)   |         |         | **83.** (1.) |            |
| *New comb.*  | 29.3 (1.1)  | 80. (3.)   | **86.** (3.) |       |         | 84. (3.)   |            |
| German-5     |             | 80. (0.)   | 82. (1.)   | 80. (1.) |        | **83.** (1.) |            |
| German-10:   |             | 80. (1.02) | 82. (0.91) |         | 79. (0.99) | **83.** (1.2) | 82. (0.96) |
| *New morph.* | 75.3 (0.7)  | 80. (1.)   | 79. (1.)   |         |         | **83.** (1.) |            |
| *New comb.*  | 24.2 (0.7)  | 79. (3.)   | **89.** (2.) |       |         | 83. (3.)   |            |

Table 3: Performance on the task of canonical segmentation for English and German (Word accuracy and standard deviation averaged over 5 splits (English-5, German-5) and over 10 splits (English-10, German-10), the rounding schemes of previously published results are applied.). RR* - neural reranker model of Kann et al. (2016). Joint* - joint transduction and segmentation model of Cotterell et al. (2016). Joint + Vec* - joint transduction and segmentation model with word embeddings (trained on Wikipedia data) of Cotterell and Schütze (2018) The subcategories of the data are described in Section 5.

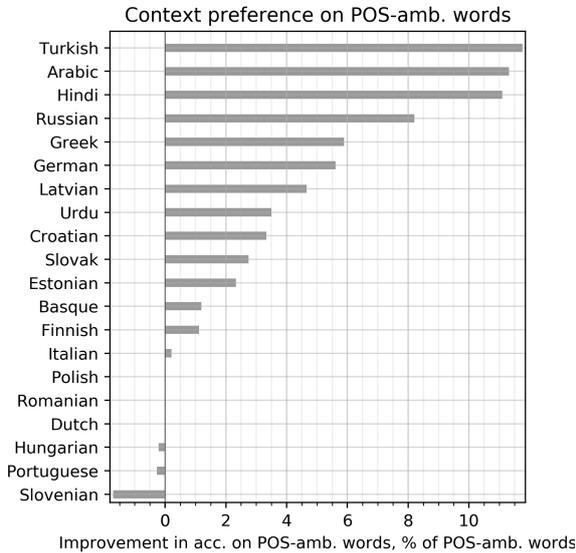

Figure 5: Context preference on POS-amb. words in the lemmatization task quantified as difference in accuracy between NMT+context+Gold POS and NMT+Gold POS models on POS-amb. category.

show that the proposed architectures improve the results of previous best-performing models on the considered data sets. In addition, we show how the proposed parts of the framework work in a complementary fashion.

Our detailed evaluation showed that including POS tags improves the performance on ambiguous words, provided a good quality POS annotation. In future work, one could investigate whether additional modifications built on top of the hierarchical bi-LSTMs encoder (e.g. conditional random field